\documentclass[nohyperref]{article}

\usepackage{microtype}
\usepackage{subfigure}

\usepackage{url, natbib, bm, amsfonts, amssymb, graphicx, anyfontsize, tabularx}
\usepackage{enumitem, booktabs}
\setcitestyle{round}

\usepackage{hyperref}

\usepackage[accepted]{icml2022}

\usepackage{amsmath}
\usepackage{amssymb}
\usepackage{mathtools}
\usepackage{amsthm}
\allowdisplaybreaks

\usepackage[capitalize,noabbrev]{cleveref}

\theoremstyle{plain}

\theoremstyle{definition}

\theoremstyle{remark}

\usepackage[textsize=tiny]{todonotes}

\icmltitlerunning{Interpretable Deep Causal Learning for Moderation Effects}

\begin{document}

\twocolumn[
\icmltitle{Interpretable Deep Causal Learning for Moderation Effects}

% It is OKAY to include author information, even for blind
% submissions: the style file will automatically remove it for you
% unless you've provided the [accepted] option to the icml2022
% package.

% List of affiliations: The first argument should be a (short)
% identifier you will use later to specify author affiliations
% Academic affiliations should list Department, University, City, Region, Country
% Industry affiliations should list Company, City, Region, Country

% You can specify symbols, otherwise they are numbered in order.
% Ideally, you should not use this facility. Affiliations will be numbered
% in order of appearance and this is the preferred way.
% \icmlsetsymbol{equal}{*}

\begin{icmlauthorlist}
\icmlauthor{Alberto Caron}{ucl,turing}
\icmlauthor{Gianluca Baio}{ucl}
\icmlauthor{Ioanna Manolopoulou}{ucl}
\end{icmlauthorlist}

\icmlaffiliation{ucl}{Department of Statistical Science, University College London, London, UK}
\icmlaffiliation{turing}{The Alan Turing Institute, London, UK}

\icmlcorrespondingauthor{Alberto Caron}{alberto.caron.19@ucl.ac.uk}

% You may provide any keywords that you
% find helpful for describing your paper; these are used to populate
% the "keywords" metadata in the PDF but will not be shown in the document
\icmlkeywords{Machine Learning, ICML}

\vskip 0.3in
]

% this must go after the closing bracket ] following \twocolumn[ ...

% This command actually creates the footnote in the first column
% listing the affiliations and the copyright notice.
% The command takes one argument, which is text to display at the start of the footnote.
% The \icmlEqualContribution command is standard text for equal contribution.
% Remove it (just {}) if you do not need this facility.

%\printAffiliationsAndNotice{}  % leave blank if no need to mention equal contribution
\printAffiliationsAndNotice{} % otherwise use the standard text.

\begin{abstract}
In this extended abstract paper, we address the problem of interpretability and targeted regularization in causal machine learning models. In particular, we focus on the problem of estimating \emph{individual causal/treatment effects} under observed confounders, which can be controlled for and moderate the effect of the treatment on the outcome of interest. Black-box ML models adjusted for the causal setting perform generally well in this task, but they lack interpretable output identifying the main drivers of treatment heterogeneity and their functional relationship. We propose a novel deep counterfactual learning architecture for estimating individual treatment effects that can simultaneously: i) convey targeted regularization on, and produce quantify uncertainty around the quantity of interest (i.e., the \emph{Conditional Average Treatment Effect}); ii) disentangle baseline \emph{prognostic} and \emph{moderating} effects of the covariates and output interpretable score functions describing their relationship with the outcome. Finally, we demonstrate the use of the method via a simple simulated experiment and a real-world application\footnote{Code for full reproducibility can be found at \url{https://github.com/albicaron/ICNN}.}. 
\end{abstract}

\section{Introduction}

In the past years, there has been a growing interest towards applying ML methods for causal inference. Disciplines such as precision medicine and socio-economic sciences inevitably call for highly personalized decision making when designing and deploying policies. Although in these fields exploration of policies in the real world through randomized experiments is costly, in order to answer counterfactual questions such as ``what would have happened if individual $i$ undertook medical treatment A instead of treatment B" one can rely on observational data, provided that the confounding factors can be controlled for. Black-box causal ML models proposed in many recent contributions perform remarkably well in the task of estimating individual counterfactual outcomes, but significantly lack interpretability, which is a key component in the design of personalized treatment rules. This is because they jointly model the outcome dependency on the covariates and on the treatment variable. Knowledge of what are the main moderating factors of a treatment can unequivocally lead to overall better policy design, as moderation effects can be leveraged to achieve higher cumulative utility when deploying the policy (e.g., by avoiding treating patient with uncertain or borderline response, better treatment allocation on budget/resources constraints,...). Another main issue of existing causal ML models, related to the one of interpretability, is carefully designed regularization \citep{wager_2020, hahn_2020, caron_2021}. Large observational studies generally include measurements on a high number of pre-treatment covariates, and disentangling prognostic\footnote{Prognostic effect is defined as the baseline effect of the covariates on the outcome, in absence of treatment.} and moderating effects allows the application of targeted regularization on both, that avoids incurring in unintended finite sample bias and large variance (see \cite{hahn_2020} for a detailed discussion on \emph{Regularization Induced Confounding} bias). This is useful in many scenarios where treatment effect is believed to be a sparser and relatively less complex function of the covariates compared to the baseline prognostic effect, so it necessitates carefully tailored regularization.

\subsection{Related Work}

Among the most influential and recent contributions on ML regression-based techniques for individualized treatment effects learning, we particularly emphasize the work of \cite{johansson_2016, shalit_2017, yao_2018} on deep learning models, \cite{vanderschaar_2017, vanderschaar_2018} on Gaussian Processes, \cite{hahn_2020, caron_2021} on Bayesian Additive Regression Trees, and finally the literature on the more general class of Meta-Learners models \cite{kunzel_2017, nie_2020}. We refer the reader to \cite{caron_2020} for a detailed review of the above methods. 

In particular we build on top of contributions by \cite{nie_2020, hahn_2020, caron_2021}, that have previously addressed the two issues of targeted regularization in causal ML. Our work proposes a new deep architecture that can separate baseline prognostic and treatment effects, and, by borrowing ideas from recent work on Neural Additive Models (NAMs) \citep{agarwal_2021}, a deep learning version of Generalized Additive Models, can output interpretable score functions describing the impact of each covariate in terms of their prognostic and treatment effects. 

\begin{figure}[t]
    \centering
    \includegraphics[scale=1]{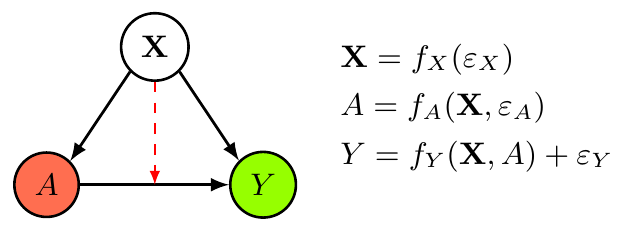}
    \caption{Causal DAG and set of structural equations describing a setting that satisfies the \emph{backdoor criterion}. The underlying assumption is that conditioning on the confounders $\bm{X}$ is sufficient to identify the causal effect $A \rightarrow Y$. Models generally assume mean-zero additive error term for the outcome equation. The red arrow in the DAG represent the moderating effect of $\bm{X}$ in the $A \rightarrow Y$ relationship.}
    \label{fig:DAG}
\end{figure}

\section{Problem Framework}

In this section we briefly introduce the main notation setup for causal effects identification and estimation under observed confounders scenarios, by utilizing the framework of Structural Causal Models (SCMs) and \emph{do}-calculus \citep{pearl_2009}. We assume we have access to data of observational nature described by the tuple $\mathbb{D}_i = \{\bm{X}_i, A_i, Y_i \} \sim p(\cdot)$, with $i \in \{1, ..., N\}$, where $\bm{X}_i \in \mathcal{X}$ is a set of covariates, $A_i \in \mathcal{A}$ a binary manipulative variable, and $Y_i \in \mathbb{R}$ is the outcome. We assume then that the causal relationships between the three variables are fully described by the SCM depicted in Figure \ref{fig:DAG}, both in the forms of causal DAG and set of structural equations. A causal DAG is a graph made of vertices and edges $(\mathcal{V}, \mathcal{E})$, where vertices represent the observational random variables, while edges represent causal functional relationships. Notice that we assume, in line with most of the literature, zero-mean additive error structure for the outcome equation. The ultimate goal is to identify and estimate the Conditional Average Treatment Effects (CATE), defined as the effect of intervening on the manipulative variable $A_i$, by setting equal to some value $a$ (or $do(A_i = a$ in the \emph{do}-calculus notation), on the outcome $Y_i$, conditional on covariates $\bm{X}_i$ (i.e., conditional on patient's characteristics, ...). In the case of binary $A_i$, CATE is defined as:
\begin{align} 
\texttt{CATE:} \,\,\,\, \tau(\bm{x}_i) & = \mathbb{E} [ Y_i \mid do(A_i = 1), \bm{X}_i = \bm{x} ]  \nonumber \\ & \,\,\, - \mathbb{E} [ Y_i \mid do(A_i = 0), \bm{X}_i = \bm{x} ] \, . \label{eq:CATE}
\end{align}
In order to identify the quantity in (\ref{eq:CATE}) we make two standard assumptions. The first assumption is that there are no unobserved confounders (\emph{unconfoundedness}) --- or equivalently in Pearl's terminology, that $\bm{X}_i$ satisfies the \emph{backdoor criterion}. The second assumption is \emph{common support}, which states that there is no deterministic selection into either of the treatment arms conditional on the covariates, or equivalently that $p(A_i = 1 | \bm{X}_i = \bm{x}) \in (0,1),~\forall i$. The latter guarantees that we could theoretically observe data points with $\bm{X}_i=\bm{x}$ in each of the two arms of $A$. Under these two assumptions, we can identify CATE $\tau(\bm{x}_i)$ in terms of observed quantities only, replacing the \emph{do}-operator in (\ref{eq:CATE}) with the factual $A_i$, by conditioning on $\bm{X}_i$:
\begin{equation*}
\mathbb{E} [ Y_i | do(A_i = a), \bm{X}_i = \bm{x} ] = \mathbb{E} [ Y_i | A_i = a, \bm{X}_i = \bm{x} ] \, .
\end{equation*}
Once CATE is identified as above, there are different ways in which it can be estimated in practice. We will briefly describe few of them in the next section.

\section{Targeted CATE estimation}

Very early works in the literature on CATE estimation proposed fitting a single model $\hat{f}_Y (\bm{X}_i, A_i)$ (S-Learners). The main drawback of S-Learners is that they are unable to account for any group-specific distributional difference, which becomes more relevant the stronger the selection bias is. Most of the subsequent contributions instead suggested splitting the sample into treatment subgroups and fit separate, arm-specific models $\hat{f}_{Y_a} (\bm{x}_i)$ (T-Learners). While T-Learners are able to account for distributional variation attributable to $A_i$, they are less sample efficient, prone to CATE overfitting and to regularization induced confounding bias \citep{kunzel_2017, hahn_2020, caron_2021}. In addition they do not produce credible intervals directly on CATE, as a CATE estimator is derived as the difference of two separate models' fit $\hat{\tau} (\bm{x}_i) = \hat{f}_1 (\bm{x}_i) - \hat{f}_0 (\bm{x}_i)$, with the induced variance being potentially very large: 
\begin{equation*}
\begin{gathered}
    \mathbb{V} \big( \hat{\tau} (\bm{x}_i) \big) = \mathbb{V} \big( \hat{f}_1 (\bm{x}_i) - \hat{f}_0 (\bm{x}_i) \big) =  \\
    = \mathbb{V} \big( \hat{f}_1 (\bm{x}_i) \big) + \mathbb{V} \big( \hat{f}_0 (\bm{x}_i) \big) - 2 Cov \big( \hat{f}_1 (\bm{x}_i), \hat{f}_0 (\bm{x}_i) \big) \, .
\end{gathered}
\end{equation*}    
Finally some of the most recent additions to the literature \citep{hahn_2020, nie_2020, caron_2021} proposed using \cite{robinson_1988} additively separable re-parametrization of the outcome function, which reads:
\begin{equation} \label{eq:rob}
    \texttt{Robinson:~} Y_i ~ = \underbrace{ \mu(\bm{X}_i) }_{\text{Prognostic Eff}} + ~~ \underbrace{ \tau (\bm{X}_i) }_{\text{CATE}} ~ A_i ~ + ~ \varepsilon_i\,,
\end{equation}
where $\mu (\bm{x}_i) = \mathbb{E} [ Y_i \mid do(A_i = 0), \bm{X}_i = \bm{x} ]$ is the prognostic effect function and $\tau (\bm{x}_i)$ is the CATE function as defined in (\ref{eq:CATE}). We assume like most contributions that $\mathbb{E} (\varepsilon_i) = 0$. The distinctive trait of Robinson's parametrization is that the outcome function explicitly includes the function of interest, i.e.\,CATE $\tau (\bm{x}_i)$, while in the usual S- or T-Learner (and subsequent variations of these) parametrizations CATE is implicitly obtained post-estimation as $\hat{\tau} (\bm{x}_i) = \hat{f}_1 (\bm{x}_i) - \hat{f}_0 (\bm{x}_i)$. This means that (\ref{eq:rob}) is able to differentiate between the baseline prognostic effect $\mu(\bm{x}_i)$ (in absence of treatment) and moderating effects embedded in the CATE function $\tau (\bm{x}_i)$, of the covariates. As a consequence, by utilizing (\ref{eq:rob}), one can convey different degree of regularization when estimating the two functions. This is particularly useful as CATE is usually believed to display simpler patterns than $\mu(\bm{x}_i)$; so by estimating it separately, one is able to apply stronger targeted regularization.

\subsection{Interpretable Causal Neural Networks}

\begin{figure}
    \centering
    \includegraphics[scale=0.96]{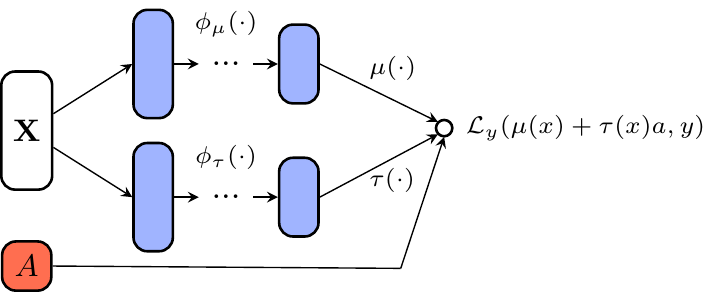}
    \caption{Intuitive TCNN structure. The deep architecture is modelled through a sample efficient, tailored loss function based on Robinson's parametrization.}
    \label{fig:TCNN}
\end{figure}

Following \cite{robinson_1988}, and the more recent work by \cite{nie_2020, hahn_2020, caron_2021}, we propose a very simple deep learning architecture for interpretable and targeted CATE estimation, based on Robinson parametrization. The architecture is made of two separable neural nets blocks that respectively learn the prognostic function $\mu(\bm{x}_i)$ and the CATE function $\tau (\bm{x}_i)$, but are ``reconnected" at the end of the pipeline to minimize a single loss function, unlike T-Learners which instead minimize separate loss functions on $f_1(\cdot)$ and $f_0(\cdot)$. Our target loss function to minimize is generally defined as follows:
\begin{equation} \label{eq:loss}
    \texttt{TCNN: } ~ \min_{\mu(\cdot), \tau(\cdot)} \mathcal{L}_y \big(\mu(\bm{x}) + \tau(\bm{x}) a, y \big)~,
\end{equation}
where $\mathcal{L}_y (\cdot)$ can be any standard loss function (e.g., MSE, negative log-likelihood,...). Through its separable block structure, the model allows the design of different NN architectures for learning $\mu(\cdot)$ and $\tau(\cdot)$ while preserving sample efficiency (i.e., avoiding sample splitting as in T-Learners), and to produce uncertainty measures around CATE $\tau(\cdot)$ directly. Thus, if $\tau(\cdot)$ is believed to display simple moderating patterns as a function of $\bm{X}_i$, a shallower NN structure with less hidden layers and units, and more aggressive regularization (e.g., higher regularization rate or dropout probabilities), can be specified, while retaining higher level of complexity in the $\mu(\cdot)$ block. We generally refer to this model as Targeted Causal Neural Network (TCNN) for simplicity from now onwards. Figure \ref{fig:TCNN} provide a simple visual representation. While in this work we focus on binary intervention variables $A_i$ for simplicity, TCNN can be easily extended to multi-category $A_i$ by adding extra blocks to the structure in Figure \ref{fig:TCNN}.

In addition to the separable structure, and in order to guarantee higher level of interpretability on prognostic and moderating factors, we also propose using a recently developed neural network version of Generalized Additive Models (GAMs), named Neural Additive Models (NAMs) \citep{agarwal_2021}, as the two $\mu(\cdot)$ and $\tau(\cdot)$ NN building blocks of TCNN. We refer to this particular version of TCNN as Interpretable Causal Neural Network (ICNN). Contrary to normal NNs, which fully ``connect" inputs to every nodes in the first hidden layer, NAMs ``connect" each single input to its own NN structure and thus outputs input-specific \emph{score functions}, that fully describe the predicted relationship between each input and the outcome. NAM's score functions have an intuitive interpretation as \emph{Shapley values} \citep{shapley_1952}: how much of an impact each input has on the final predicted outcome. The structure of the loss function (\ref{eq:loss}) in ICNN thus becomes additive also in the $P$ covariate-specific $\mu_j(\cdot)$ and $\tau_j(\cdot)$ functions:
\begin{equation*}
    \texttt{ICNN: } \min_{\mu(\cdot), \tau(\cdot)} \mathcal{L}_y \Big(\sum^P_{j=1} \mu_j(x_j) + \sum^P_{j=1} \tau_j(x_j) a, \, y \Big)~,
\end{equation*}
where the single $\mu_j(x_j)$ score function represents the Shapley value in terms of prognostic effect of covariate $x_j$, while $\tau_j(x_j)$ its Shapley value in terms of moderating effect. Hence, the NAM architecture in ICNN allows us to estimate the impact of each covariates as a prognostic and moderating factor and quantify the uncertainty around them as well. Under ICNN, the outcome function thus becomes twice additively separable, as:
\begin{equation}
    Y_i = \sum^P_{j=1} \mu_j(x_{i,j}) + \sum^P_{j=1} \tau_j(x_{i,j}) A_i + \varepsilon_i \,,
\end{equation}
where $i \in \{1,...,N\}$ and $j \in \{1,...,P\}$. Naturally, the downside of NAMs is that they might miss out on interaction terms among the covariates. These could possibly be constructed and added manually as additional inputs, although this is not particularly convenient nor computationally ideal.

\subsection{Links to Previous Work}

We conclude the section by highlighting similarities and differences between TCNN (and ICNN) and other popular methods employing Robinson's parametrization. Differently than R-Learner \citep{nie_2020}, TCNN is not a multi-step plug-in (and cross-validated) estimator and does not envisage the use of propensity score. Instead, similarly to Bayesian Causal Forest (BCF) \citep{hahn_2020, caron_2021}, estimation in TCNN is carried out in a single, more sample efficient step, although BCF is inherently Bayesian and relatively computationally intensive. To obtain better coverage properties in terms of uncertainty quantification in both TCNN and ICNN, we implement the MC dropout technique \citep{gal_2016} in both $\mu(\cdot)$ and $\tau(\cdot)$ blocks to perform approximate Bayesian inference, that is, we re-sample multiple times from the NN model with dropout layers to build an approximate posterior predictive distribution. This produces credible intervals around CATE estimates $\tau(\cdot)$ in a very straightforward way, and, in ICNN specifically, credible intervals around each inputs' score function, as we will show in the experimental section.

\begin{table}[t]
    \centering
    \begin{tabular}{l | c | c}
        \multicolumn{1}{c|}{\textbf{Model}} & \textbf{Train} $\bm{\sqrt{\text{\textbf{PEHE}}_\tau}}$ &  \textbf{Test} $\bm{\sqrt{\text{\textbf{PEHE}}_\tau}}$ \\
        \midrule
        S-NN  & 1.046 $\pm$ 0.007  & 1.076 $\pm$ 0.007 \\
        T-NN  &  1.021 $\pm$ 0.002 & 1.074 $\pm$ 0.002 \\
        \midrule
        R-CF  & 1.467 $\pm$ 0.002 & 1.494 $\pm$ 0.002 \\
        R-NN  & 0.706 $\pm$ 0.003 & 0.712 $\pm$ 0.003 \\
        R-NAM & 0.787 $\pm$ 0.002 & 0.787 $\pm$ 0.002 \\
        \midrule
        TCNN  & \textbf{0.361 $\pm$ 0.001} & \textbf{0.362 $\pm$ 0.001} \\
        ICNN  & \textbf{0.328 $\pm$ 0.001} &  \textbf{0.331 $\pm$ 0.001}
    \end{tabular}
    \caption{Performance on simulated experiment, measured as 70\%-30\% train-test set $\bm{\sqrt{\text{PEHE}_\tau}}$. Bold indicates better performance.}
    \label{tab:result}
\end{table}

\section{Experiments} \label{sec:exp}

We hereby present results from a simple simulated experiment on CATE estimation, to compare TCNN and ICNN performance against some of the state of the art methods. In addition, we demonstrate how ICNN with MC dropout in particular can be employed to produce highly interpretable score function measures, fully describing the estimated moderating effects of the covariates $\bm{x}_i$ in $\tau(\cdot)$, and uncertainty around them. For performance comparison we rely on the root \emph{Precision in Estimating Heterogeneous Treatment Effects} (PEHE) metric \citep{hill_2011}, defined as:
\begin{equation} 
    \sqrt{\text{PEHE}_\tau} = \sqrt{ \mathbb{E} \big[ (\hat{\tau}_i (\bm{x}_i) - \tau_i (\bm{x}_i) )^2 \big] } \, ,
\end{equation}
and the list of models we compare include: S-Learner version of NNs (S-NN); T-Learner version of NNs (T-NN); Causal Forest \citep{athey_2019}, a particular type of R-Learner (R-CF); a ``unique-block", fully connected NN that uses Robinson's parametrization minimizing the loss function in (\ref{eq:loss}) (R-NN); a ``unique-block" NAM, again minimizing the loss function in (\ref{eq:loss}) (R-NAM); our TCNN with fully connected NN blocks; and ICNN. S-NN, T-NN and R-NN all feature two [50, 50] hidden layers. R-NAM features two [20, 20] hidden layers for each input. TCNN features two [50, 50] hidden layers in the $\mu(\cdot)$ block, and one [20] hidden layer in the $\tau(\cdot)$ block. ICNN features two [20, 20] hidden layers in the $\mu(\cdot)$ block, and one [50] hidden layer in the $\tau(\cdot)$ block, for each input.

\begin{figure}
    \centering
    \includegraphics[scale=0.5]{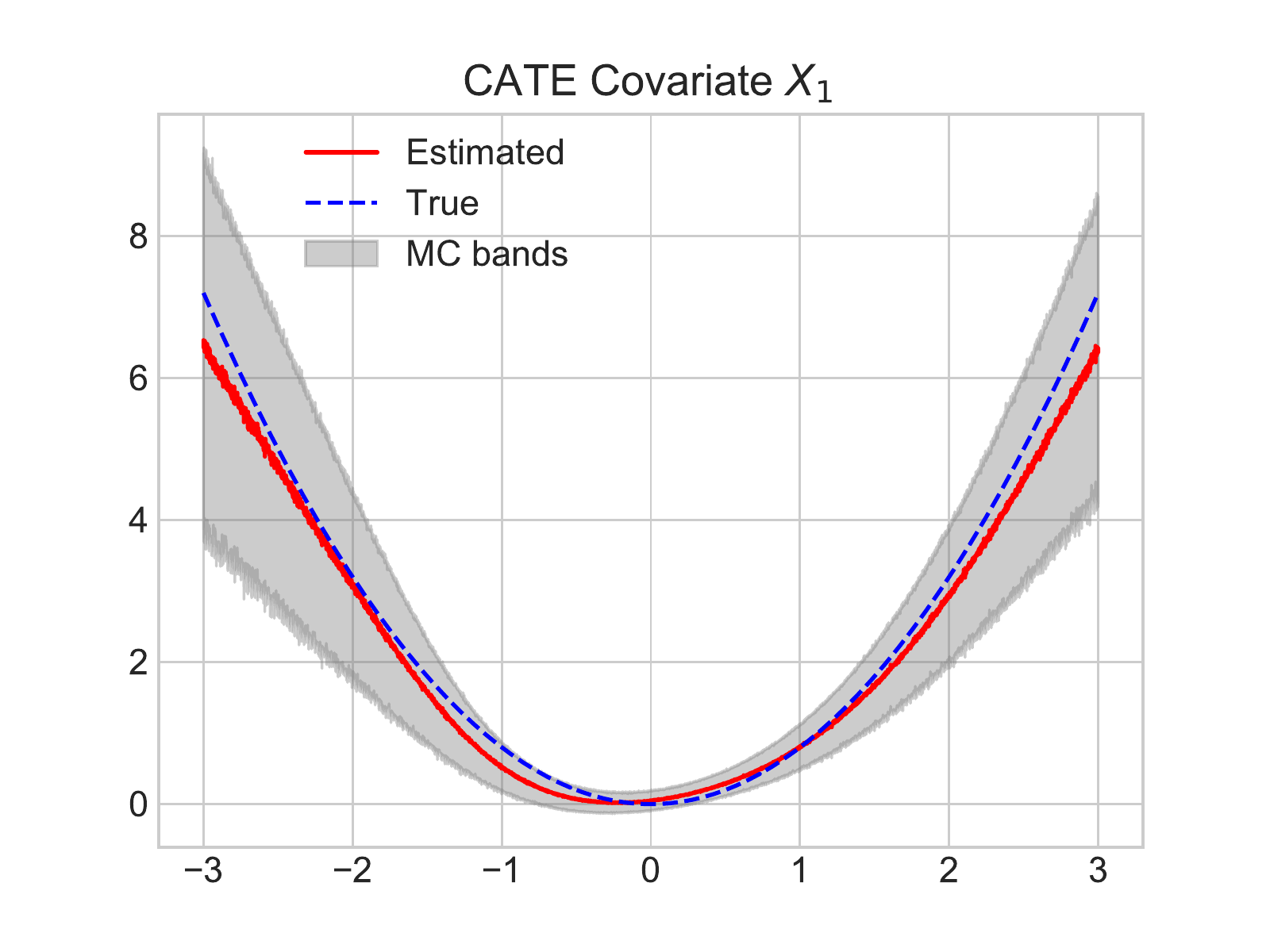}
    \caption{Score function output from ICNN model relative to covariate $X_1$, depicting its moderating effect on CATE, plus MC dropout generated credible intervals.}
    \label{fig:ICNN}
\end{figure}

We simulate $N=2000$ data points on $P=10$ correlated covariates, with binary $A_i$ and continuous $Y_i$. The experiment was run for $B=100$ replications and results on 70\%-30\% train-test sets average $\sqrt{\text{PEHE}_\tau}$, plus 95\% Monte Carlo errors, can be found in Table \ref{tab:result}. The full description of the data generating process utilized for this simulated experiment can be found in the appendix Section \ref{sec:appA}. NN models minimizing the Robinson loss function in (\ref{eq:loss}) perform considerably better than S- and T-Learner baselines on this particular example, especially TCNN and ICNN that present the additional advantage of conveying targeted regularization. Considering the ICNN model only, we can then access the score functions on the $\tau(\cdot)$ NAM block that describe the moderating effects of the covariates $\bm{x}_i$. In particular in Figure \ref{fig:ICNN} we plot the score function of the first covariate $X_{i, 1}$ on CATE $\tau(\cdot)$, plus the approximate Bayesian credible intervals generated through MC dropout resampling \citep{gal_2016}. In this specific simulated example, CATE function is generated as $\tau(\bm{x}_i) = 3 + 0.8 X_{i, 1}^2$. So only $X_{i, 1}$, out of all $P=10$ covariates, drives the simple heterogeneity patterns in treatment response across individuals, in a quadratic form. As Figure \ref{fig:ICNN} shows, ICNN is able in this example to learn a score function that very closely approximates the underlying true relationship $0.8 X_{i, 1}^2$, and quantifies uncertainty around it. Naturally, in a different simulated setup with strong interaction terms among the covariates, performance of ICNN would inevitably deteriorate compared to the other versions of NN and models considered here. Thus, performance and interpretability in this type of scenario would certainly constitute a trade-off.

\begin{figure*}[t]
    \centering
    \includegraphics[scale=0.34]{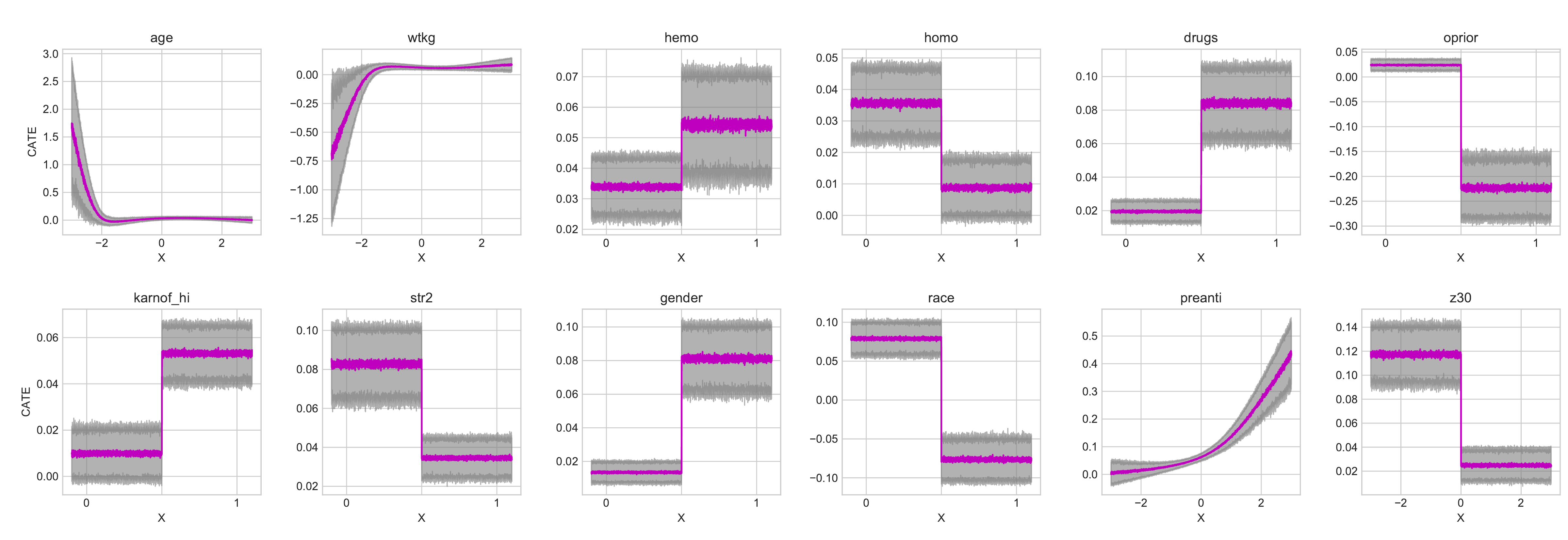}
    \caption{Score functions (or Shapley values) with associated MC dropout bands describing moderation effects of each covariate on estimated CATE: $\tau_j (x_j), \, \forall j \in \{1,...,P\}$.}
    \label{fig:ACTG}
\end{figure*}

\subsection{Real-World Example: the ACTG-175 data}  \label{sec:ACTG}

Finally, we briefly demonstrate the use of ICNN on a real-world example. Although the focus of the paper so far has been on observational type of studies, we will analyze data from a randomized experiment to show that the methods introduced naturally extend to this setting as well, with the non-negligible additional benefit that both \emph{unconfoundedness} and \emph{common support} assumptions hold by construction (i.e., no ``causal" arrow going from $\bm{X} \rightarrow A$ in Figure \ref{fig:DAG} DAG). The data we use are taken from the ACTG-175 study, a randomized controlled trial comparing standard mono-therapy against a combination of therapies in the treatment of HIV-1-infected patients with CD4 cell counts between 200 and 500. Details of the design can be found in the original contribution by \citet{hammer_1996}. The dataset features $N=2139$ observations and $P=12$ covariates $\bm{X}$ (which are listed in the appendix section below), a binary treatment $A$ (mono-therapy VS multi-therapy) and a continuous outcome $Y$ (difference in CD4 cell counts between baseline and after 20$\pm$5 weeks after undertaking the treatment --- this is done in order to take into account any individual unobserved time pattern in the CD4 cell count). 

The aim is to investigate the moderation effects of the covariates in terms of heterogeneity of treatment across patients. In order to do so, we run ICNN and obtain the estimated score functions, together with approximated Bayesian MC dropout bands, for each covariate $X_j$, and we report these in Figure \ref{fig:ACTG}. The results generally suggest a good degree of treatement heterogeneity, with most of the covariates playing a significant moderating role. 

\section{Conclusion}

In this extended abstract paper, we have addressed the issue of interpretability and targeted regularization in causal machine learning models for the estimation of heterogeneous/individual treatment effects. In particular we have proposed a novel deep learning architecture (TCNN) that is able to convey regularization and quantify uncertainty when learning the CATE function, and, in its interpretable version (ICNN), to output interpretable score function describing the estimated prognostic and moderation effects of the covariates $\bm{X}_i$. We have benchmarked TCNN and ICNN by comparing their performance against some of the popular methods for CATE estimation on a simple simulated experiment, where we have also illustrated how score functions are very intuitive and interpretable measures for moderation effects analysis. Finally, we have demonstrated the use of ICNN on a real-world dataset based on the ACTG-175 study \citep{hammer_1996}.

% In the unusual situation where you want a paper to appear in the
% references without citing it in the main text, use \nocite
\nocite{*}

\bibliography{Refs}

\begin{thebibliography}{36}
\providecommand{\natexlab}[1]{#1}
\providecommand{\url}[1]{\texttt{#1}}
\expandafter\ifx\csname urlstyle\endcsname\relax
  \providecommand{\doi}[1]{doi: #1}\else
  \providecommand{\doi}{doi: \begingroup \urlstyle{rm}\Url}\fi

\bibitem[Abdar et~al.(2021)Abdar, Pourpanah, Hussain, Rezazadegan, Liu,
  Ghavamzadeh, Fieguth, Cao, Khosravi, Acharya, Makarenkov, and
  Nahavandi]{ABDAR2021243}
Abdar, M., Pourpanah, F., Hussain, S., Rezazadegan, D., Liu, L., Ghavamzadeh,
  M., Fieguth, P., Cao, X., Khosravi, A., Acharya, U.~R., Makarenkov, V., and
  Nahavandi, S.
\newblock A review of uncertainty quantification in deep learning: Techniques,
  applications and challenges.
\newblock \emph{Information Fusion}, 76:\penalty0 243--297, 2021.

\bibitem[Agarwal et~al.(2021)Agarwal, Melnick, Frosst, Zhang, Lengerich,
  Caruana, and Hinton]{agarwal_2021}
Agarwal, R., Melnick, L., Frosst, N., Zhang, X., Lengerich, B., Caruana, R.,
  and Hinton, G.~E.
\newblock Neural additive models: Interpretable machine learning with neural
  nets.
\newblock In \emph{Proceedings of the 35th International Conference on Neural
  Information Processing Systems}, volume~34, pp.\  4699--4711, 2021.

\bibitem[Alaa \& van~der Schaar(2018)Alaa and van~der
  Schaar]{vanderschaar_2018}
Alaa, A. and van~der Schaar, M.
\newblock Limits of estimating heterogeneous treatment effects: Guidelines for
  practical algorithm design.
\newblock In \emph{Proceedings of the 35th International Conference on Machine
  Learning}, pp.\  129--138, 2018.

\bibitem[Alaa \& van~der Schaar(2017)Alaa and van~der
  Schaar]{vanderschaar_2017}
Alaa, A.~M. and van~der Schaar, M.
\newblock Bayesian inference of individualized treatment effects using
  multi-task {G}aussian {P}rocesses.
\newblock In \emph{Proceedings of the 31st International Conference on Neural
  Information Processing Systems}, NIPS'17, pp.\  3427–3435, 2017.

\bibitem[Athey \& Wager(2021)Athey and Wager]{athey_2020}
Athey, S. and Wager, S.
\newblock {Policy Learning With Observational Data}.
\newblock \emph{Econometrica}, 89\penalty0 (1):\penalty0 133--161, January
  2021.

\bibitem[Blundell et~al.(2015)Blundell, Cornebise, Kavukcuoglu, and
  Wierstra]{10.5555/3045118.3045290}
Blundell, C., Cornebise, J., Kavukcuoglu, K., and Wierstra, D.
\newblock Weight uncertainty in neural networks.
\newblock In \emph{Proceedings of the 32nd International Conference on
  International Conference on Machine Learning - Volume 37}, pp.\  1613–1622,
  2015.

\bibitem[Caron et~al.(2022{\natexlab{a}})Caron, Baio, and
  Manolopoulou]{caron_2020}
Caron, A., Baio, G., and Manolopoulou, I.
\newblock Estimating individual treatment effects using non-parametric
  regression models: A review.
\newblock \emph{Journal of the Royal Statistical Society: Series A (Statistics
  in Society)}, pp.\  1--35, 2022{\natexlab{a}}.

\bibitem[Caron et~al.(2022{\natexlab{b}})Caron, Baio, and
  Manolopoulou]{caron_2021}
Caron, A., Baio, G., and Manolopoulou, I.
\newblock Shrinkage bayesian causal forests for heterogeneous treatment effects
  estimation.
\newblock \emph{Journal of Computational and Graphical Statistics}, pp.\
  1--13, 2022{\natexlab{b}}.

\bibitem[Chernozhukov et~al.(2018)Chernozhukov, Chetverikov, Demirer, Duflo,
  Hansen, Newey, and Robins]{duflo_2018}
Chernozhukov, V., Chetverikov, D., Demirer, M., Duflo, E., Hansen, C., Newey,
  W., and Robins, J.
\newblock Double/debiased machine learning for treatment and structural
  parameters.
\newblock \emph{The Econometrics Journal}, 21\penalty0 (1):\penalty0 C1--C68,
  2018.

\bibitem[Chipman et~al.(2010)Chipman, George, and McCulloch]{chipman_2010}
Chipman, H.~A., George, E.~I., and McCulloch, R.~E.
\newblock {BART}: Bayesian additive regression trees.
\newblock \emph{Ann. Appl. Stat.}, 4\penalty0 (1):\penalty0 266--298, 03 2010.

\bibitem[Gal \& Ghahramani(2016)Gal and Ghahramani]{gal_2016}
Gal, Y. and Ghahramani, Z.
\newblock Dropout as a bayesian approximation: Representing model uncertainty
  in deep learning.
\newblock In \emph{Proceedings of The 33rd International Conference on Machine
  Learning}, volume~48, pp.\  1050--1059, 2016.

\bibitem[Hahn et~al.(2018)Hahn, Carvalho, Puelz, and He]{hahn_2018}
Hahn, P.~R., Carvalho, C.~M., Puelz, D., and He, J.
\newblock Regularization and confounding in linear regression for treatment
  effect estimation.
\newblock \emph{Bayesian Anal.}, 13\penalty0 (1):\penalty0 163--182, 03 2018.

\bibitem[Hahn et~al.(2020)Hahn, Murray, and Carvalho]{hahn_2020}
Hahn, P.~R., Murray, J.~S., and Carvalho, C.~M.
\newblock {Bayesian Regression Tree Models for Causal Inference:
  Regularization, Confounding, and Heterogeneous Effects}.
\newblock \emph{Bayesian Analysis}, 15\penalty0 (3):\penalty0 965 -- 1056,
  2020.

\bibitem[Hammer et~al.(1996)Hammer, Katzenstein, Hughes, Gundacker, Schooley,
  Haubrich, Henry, Lederman, Phair, Niu, Hirsch, and Merigan]{hammer_1996}
Hammer, S.~M., Katzenstein, D.~A., Hughes, M.~D., Gundacker, H., Schooley,
  R.~T., Haubrich, R.~H., Henry, W.~K., Lederman, M.~M., Phair, J.~P., Niu, M.,
  Hirsch, M.~S., and Merigan, T.~C.
\newblock A trial comparing nucleoside monotherapy with combination therapy in
  hiv-infected adults with {CD4} cell counts from 200 to 500 per cubic
  millimeter.
\newblock \emph{N. Engl. J. Med.}, 335:\penalty0 1081--1090, 1996.

\bibitem[Hartford et~al.(2017)Hartford, Lewis, Leyton-Brown, and
  Taddy]{hartford_2017}
Hartford, J., Lewis, G., Leyton-Brown, K., and Taddy, M.
\newblock Deep {IV}: A flexible approach for counterfactual prediction.
\newblock In \emph{Proceedings of the 34th International Conference on Machine
  Learning}, volume~70, pp.\  1414--1423, 2017.

\bibitem[Hill(2011)]{hill_2011}
Hill, J.~L.
\newblock Bayesian nonparametric modeling for causal inference.
\newblock \emph{Journal of Computational and Graphical Statistics}, 20\penalty0
  (1):\penalty0 217--240, 2011.

\bibitem[Hodson(2016)]{precisionmed}
Hodson, R.
\newblock Precision medicine.
\newblock \emph{Nature}, 547\penalty0 (7619), 2016.

\bibitem[Horvitz \& Thompson(1952)Horvitz and Thompson]{horvitz_1952}
Horvitz, D.~G. and Thompson, D.~J.
\newblock A generalization of sampling without replacement from a finite
  universe.
\newblock \emph{Journal of the American Statistical Association}, 47\penalty0
  (260):\penalty0 663--685, 1952.

\bibitem[Imbens \& Rubin(2015)Imbens and Rubin]{imbens_rubin_2015}
Imbens, G.~W. and Rubin, D.~B.
\newblock \emph{Causal Inference for Statistics, Social, and Biomedical
  Sciences: An Introduction}.
\newblock Cambridge University Press, 2015.

\bibitem[Johansson et~al.(2016)Johansson, Shalit, and Sontag]{johansson_2016}
Johansson, F., Shalit, U., and Sontag, D.
\newblock Learning representations for counterfactual inference.
\newblock In \emph{Proceedings of The 33rd International Conference on Machine
  Learning}, volume~48, pp.\  3020--3029, 2016.

\bibitem[Kaddour et~al.(2021)Kaddour, Zhu, Liu, Kusner, and Silva]{jean_2021}
Kaddour, J., Zhu, Y., Liu, Q., Kusner, M.~J., and Silva, R.
\newblock Causal effect inference for structured treatments.
\newblock In \emph{Proceedings of the 35th International Conference on Neural
  Information Processing Systems}, volume~34, pp.\  24841--24854, 2021.

\bibitem[Kitagawa \& Tetenov(2018)Kitagawa and Tetenov]{kitagawa_2018}
Kitagawa, T. and Tetenov, A.
\newblock Who should be treated? empirical welfare maximization methods for
  treatment choice.
\newblock \emph{Econometrica}, 86\penalty0 (2):\penalty0 591--616, 2018.

\bibitem[Künzel et~al.(2017)Künzel, Sekhon, Bickel, and Yu]{kunzel_2017}
Künzel, S., Sekhon, J., Bickel, P., and Yu, B.
\newblock Meta-learners for estimating heterogeneous treatment effects using
  machine learning.
\newblock \emph{Proceedings of the National Academy of Sciences}, 116, 06 2017.

\bibitem[Lakshminarayanan et~al.(2017)Lakshminarayanan, Pritzel, and
  Blundell]{deepensembles_2017}
Lakshminarayanan, B., Pritzel, A., and Blundell, C.
\newblock Simple and scalable predictive uncertainty estimation using deep
  ensembles.
\newblock In \emph{Proceedings of the 31st International Conference on Neural
  Information Processing Systems}, pp.\  6405–6416. Curran Associates Inc.,
  2017.

\bibitem[Nie \& Wager(2020)Nie and Wager]{wager_2020}
Nie, X. and Wager, S.
\newblock {Quasi-oracle estimation of heterogeneous treatment effects}.
\newblock \emph{Biometrika}, 108\penalty0 (2):\penalty0 299--319, 09 2020.

\bibitem[Nie et~al.(2020)Nie, Brunskill, and Wager]{nie_2020}
Nie, X., Brunskill, E., and Wager, S.
\newblock Learning when-to-treat policies.
\newblock \emph{Journal of the American Statistical Association}, 0\penalty0
  (ja):\penalty0 1--58, 2020.

\bibitem[Pearce et~al.(2020)Pearce, Leibfried, and
  Brintrup]{pmlr-v108-pearce20a}
Pearce, T., Leibfried, F., and Brintrup, A.
\newblock Uncertainty in neural networks: Approximately bayesian ensembling.
\newblock In \emph{Proceedings of the Twenty Third International Conference on
  Artificial Intelligence and Statistics}, volume 108, pp.\  234--244, 2020.

\bibitem[Pearl(2009)]{pearl_2009}
Pearl, J.
\newblock \emph{Causality: Models, Reasoning and Inference}.
\newblock Cambridge University Press, USA, 2nd edition, 2009.
\newblock ISBN 052189560X.

\bibitem[Peters et~al.(2017)Peters, Janzing, and Schlkopf]{peters_2017}
Peters, J., Janzing, D., and Schlkopf, B.
\newblock \emph{Elements of Causal Inference: Foundations and Learning
  Algorithms}.
\newblock The MIT Press, 2017.

\bibitem[Robinson(1988)]{robinson_1988}
Robinson, P.~M.
\newblock Root-n-consistent semiparametric regression.
\newblock \emph{Econometrica}, 56\penalty0 (4):\penalty0 931--954, 1988.

\bibitem[Rubin(1978)]{rubin1978}
Rubin, D.~B.
\newblock Bayesian inference for causal effects: The role of randomization.
\newblock \emph{Ann. Statist.}, 6\penalty0 (1):\penalty0 34--58, 01 1978.

\bibitem[Shalit et~al.(2017)Shalit, Johansson, and Sontag]{shalit_2017}
Shalit, U., Johansson, F.~D., and Sontag, D.
\newblock Estimating individual treatment effect: Generalization bounds and
  algorithms.
\newblock In \emph{Proceedings of the 34th International Conference on Machine
  Learning - Volume 70}, volume~70, pp.\  3076–3085, 2017.

\bibitem[Shapley(1953)]{shapley_1952}
Shapley, L.~S.
\newblock A value for n-person games.
\newblock In \emph{Contributions to the Theory of Games II}, pp.\  307--317.
  Princeton University Press, 1953.

\bibitem[Wager \& Athey(2018)Wager and Athey]{athey_2019}
Wager, S. and Athey, S.
\newblock Estimation and inference of heterogeneous treatment effects using
  random forests.
\newblock \emph{Journal of the American Statistical Association}, 113\penalty0
  (523):\penalty0 1228--1242, 2018.

\bibitem[Yao et~al.(2018)Yao, Li, Li, Huai, Gao, and Zhang]{yao_2018}
Yao, L., Li, S., Li, Y., Huai, M., Gao, J., and Zhang, A.
\newblock Representation learning for treatment effect estimation from
  observational data.
\newblock In \emph{Advances in Neural Information Processing Systems 31}, pp.\
  2633--2643, 2018.

\bibitem[Zhang et~al.(2012)Zhang, Tsiatis, Laber, and Davidian]{zhang_2012}
Zhang, B., Tsiatis, A.~A., Laber, E.~B., and Davidian, M.
\newblock A robust method for estimating optimal treatment regimes.
\newblock \emph{Biometrics}, 68\penalty0 (4):\penalty0 1010--1018, 2012.

\end{thebibliography}
\bibliographystyle{icml2022}

%%%%%%%%%%%%%%%%%%%%%%%%%%%%%%%%%%%%%%%%%%%%%%%%%%%%%%%%%%%%%%%%%%%%%%%%%%%%%%%
%%%%%%%%%%%%%%%%%%%%%%%%%%%%%%%%%%%%%%%%%%%%%%%%%%%%%%%%%%%%%%%%%%%%%%%%%%%%%%%
% APPENDIX
%%%%%%%%%%%%%%%%%%%%%%%%%%%%%%%%%%%%%%%%%%%%%%%%%%%%%%%%%%%%%%%%%%%%%%%%%%%%%%%
%%%%%%%%%%%%%%%%%%%%%%%%%%%%%%%%%%%%%%%%%%%%%%%%%%%%%%%%%%%%%%%%%%%%%%%%%%%%%%%
\newpage
\appendix
\onecolumn

\section{Data Generating Process} \label{sec:appA}

In this appendix section we briefly describe the data generating process utilized for the simulated experiment in Section \ref{sec:exp}. We generated $N=2000$ data points on $P=10$ correlated covariates, of which 5 continuous and 5 binary, drawn from a Gaussian Copula $C^{\text{Gauss}}_{\Theta} (u) = \Phi_{\Theta} \big( \Phi^{-1}(u_1), \dots , \Phi^{-1}(u_P) \big) $, where the covariance matrix is such that $\Theta_{jk} = 0.1^{|j - k|} + 0.1\mathbb{I}(j \neq k) $. The data generating process is fully described by the following quantities: 
\begin{equation} \label{eq:tarselstudy}
\begin{gathered}
\mu (\bm{x}_i) = ~ 6 + 0.3 \exp(X_{i,1}) + 1 X^2_{i,2} + 1.5 | X_{i, 3} | + 0.8 X_{i,4} ~ , \\
\tau (\bm{x}_i) = ~ 3 + 0.8 X^2_{i,1} ~ , \\
\pi (\bm{x}_i) = ~ \Lambda \left( -1.5 + 0.5 X_{i, 1} + \frac{\nu_i}{10} \right) ~ ,  \\
A_i \sim  ~  \text{Bernoulli} \big( \pi (\bm{x}_i) \big) ~ ,  \\
Y_i =  ~  \mu (\bm{x}_i) + \tau (\bm{x}_i) A_i + \varepsilon_i ~ , \quad \text{where} \quad \varepsilon_i \sim \mathcal{N} (0, \sigma^2) ~ ,
\end{gathered}
\end{equation}
where: $\Lambda(\cdot)$ is the logistic cumulative distribution function; the error's standard deviation is $\sigma^2 = 0.5$; and $\nu_i \sim \text{Uniform} (0, 1)$. More details on the DGP and the models employed can be found at \url{https://github.com/albicaron/ICNN}, for full reproducibility.

\section{The ACTG-175 Trial Data} \label{sec:appenB}

In Table here below we report the description of the 12 covariates utilized in the analysis in Section \ref{sec:ACTG}.

\begin{table}[h]
		\caption{ACTG-175 data covariates $\bm{X}$}
		\centering \footnotesize
		\begin{tabularx}{10.5cm}{l | X}
			
			\textbf{Variable}  &  \textbf{Description} \\
			\midrule
			
			\textit{age} &  Numeric	\\	
			\textit{wtkg} &	Numeric	\\	
			\textit{hemo} &	 Binary (hemophilia = 1)	\\	
			\textit{homo} &	 Binary (homosexual = 1)	\\
			\textit{drugs} & Binary (intravenous drug use = 1)	\\
			\textit{oprior} & Binary (non-zidovudine antiretroviral therapy prior to initiation of study treatment = 1) \\
			\textit{z30} &	Binary (zidovudine use in the 30 days prior to treatment initiation = 1)	\\
			\textit{preanti} &	Numeric (number of days of previously received antiretroviral therapy)	\\	
			\textit{race} &	 Binary	\\	
			\textit{gender} &	Binary 	\\
			\textit{str2} &	 Binary: antiretroviral history (0 = naive, 1 = experienced)   \\
			\textit{karnof\_hi} &	Binary: Karnofsky score (0 = $<100$, 1 = $100$)	
			\label{AIDSvar}
		\end{tabularx}
	\end{table}

\end{document}